\documentclass[letterpaper]{article} 
\usepackage[preprint]{aaai2027}  
\usepackage[hyphens]{url}  
\usepackage{graphicx} 
\usepackage{natbib}  
\usepackage{caption} 
\usepackage{amsmath}
\usepackage{amssymb}
\usepackage{algorithm}
\usepackage{algorithmic}
\usepackage{multirow}
\usepackage{array}
\usepackage{booktabs}
\usepackage{pifont}

\newcolumntype{C}[1]{>{\centering\arraybackslash}m{#1}}
\newcommand{\cmark}{\ding{51}}
\newcommand{\xmark}{\ding{55}}

\title{MotionPhys: Detecting AI-Generated Videos via Physical Consistency of Optical-Flow Trajectories}

\author{
Haojin He\textsuperscript{\rm 1,2},
Hao Tan\textsuperscript{\rm 1},
Zichang Tan\textsuperscript{\rm 3},
Ajian Liu\textsuperscript{\rm 1},
Jun Wan\textsuperscript{\rm 1,2}
}

\affiliations{
\textsuperscript{\rm 1}State Key Laboratory of Multimodal Artificial Intelligence Systems,\\
Institute of Automation, Chinese Academy of Sciences, Beijing 100190, China\\
\textsuperscript{\rm 2}School of Artificial Intelligence, University of Chinese Academy of Sciences, Beijing 100049, China\\
\textsuperscript{\rm 3}Sangfor Technologies Inc., Shenzhen 518055, China\\
hehaojin2026@ia.ac.cn,
tanhao2023@ia.ac.cn,
tanzichang@foxmail.com,\\
ajian.liu@ia.ac.cn,
jun.wan@ia.ac.cn
}

\begin{document}

\maketitle

\begin{abstract}
Modern AI video generation models can produce videos with high visual fidelity and seemingly smooth temporal transitions. However, visual realism does not necessarily imply physical motion consistency. Existing generative models mainly optimize distribution matching in pixel or latent spaces, without explicitly enforcing real-world constraints such as inertia, continuous forces, and trajectory geometry. Our experiments show that AI-generated videos remain visually plausible over short sequences of consecutive frames, yet fail to preserve physical motion consistency throughout a complete object action, resulting in systematic statistical discrepancies in their motion trajectories. Based on this observation, we introduce \textbf{MotionPhys}, a lightweight and interpretable framework that treats sparse motion trajectories as physical evidence rather than relying on appearance artifacts or generator-specific traces. By modeling the geometric evolution of trajectories across multiple temporal scales, MotionPhys reveals subtle motion inconsistencies that are difficult to capture with conventional visual cues and transforms them into a compact representation for efficient detection. Experiments on multiple datasets show that MotionPhys can effectively detect physical inconsistencies in generated videos and generalizes well across different video generators.
\end{abstract}


\section{Introduction}
In recent years, AI video generation has developed rapidly. Models built upon diffusion models and transformers are now able to generate videos with complex scenes and high visual continuity~\cite{khachatryan2023text2video, zhang2023i2vgen, blattmann2023stable, yang2025cogvideox}. Existing detection methods have attempted to distinguish real and AI-generated videos from different perspectives. Some methods employ 3D CNNs, video transformers, or Mamba-based networks to learn to identify generative artifacts in RGB videos and then perform classification~\cite{chen2026demamba,bai2024ai}. Other methods detect or amplify subtle traces of the generation process in noise patterns, frequency bands, bit planes, or other low-level representations.~\cite{tan2024rethinking,cheng2026revealing,corvi2026seeing}. In essence, these methods mainly detect artifacts introduced during synthesis. However, such methods typically rely on frame-level differences, depth maps, or representations extracted by visual feature extractors to find statistical shifts. Although they can determine whether abnormal changes occur between adjacent frames, they may also cause the detector to overfit to a specific data distribution or generator-specific artifact, rather than capturing the intrinsic defect of AI-generated videos.

\begin{figure}[!t]
\centering
\includegraphics[width=0.98\columnwidth]{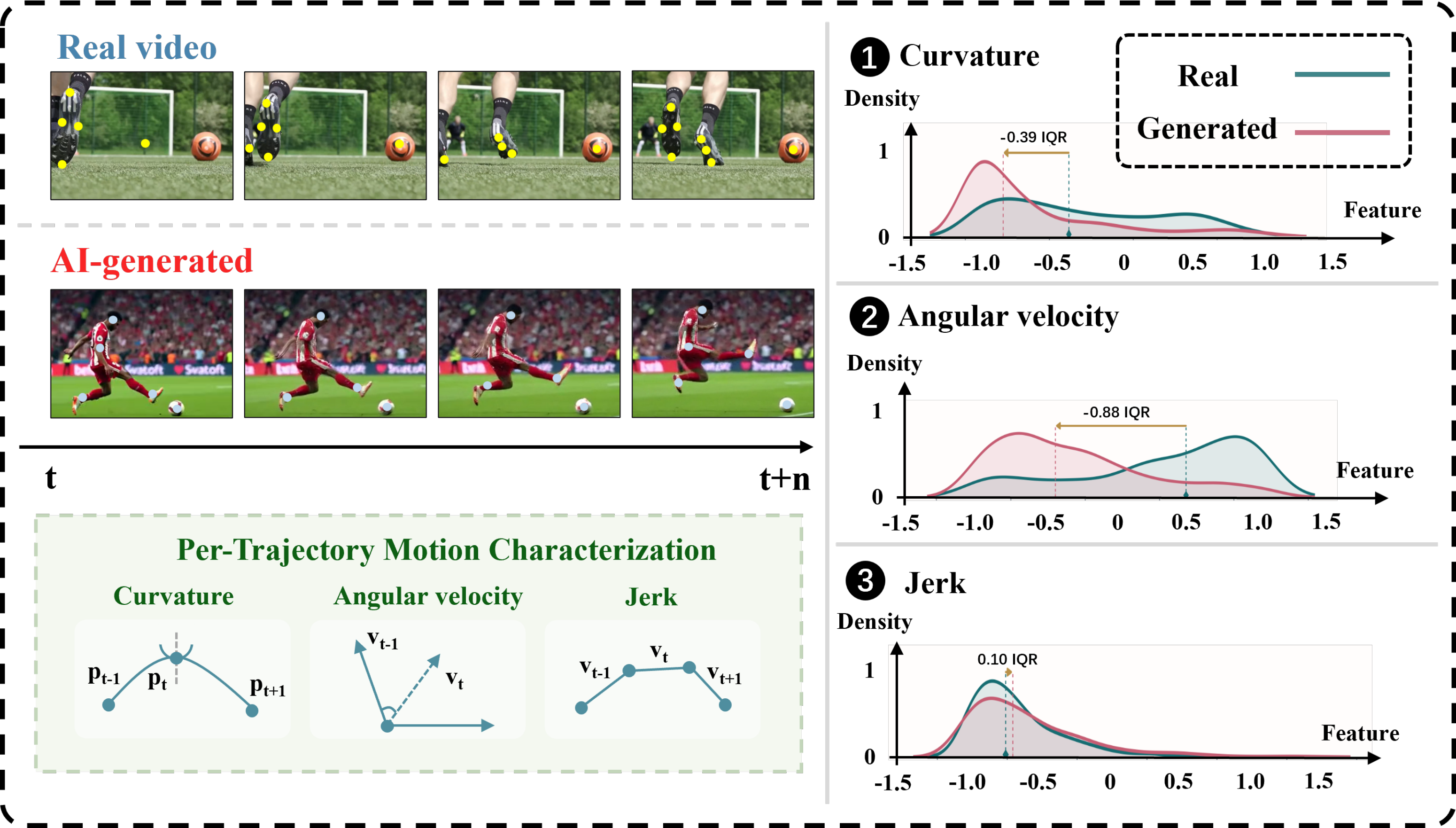}
\caption{
The left panel shows representative trajectories from a soccer-kick sequence, while the right panel presents distributions of trajectory statistics computed from real and AI-generated videos in GenBuster++. Despite similar short-term visual appearance, the two classes exhibit systematic differences in motion coordination, directional variation, and high-order temporal stability.
}
\label{fig:motivation}
\end{figure}

As video generation models continue to improve, overt synthesis artifacts are becoming increasingly scarce, making detectors that rely on appearance anomalies or generator-specific traces more likely to fail on newly developed generators. Rather than continuing to fit such purely visual cues, we turn to the physical coherence of object motion and ask whether it can remain plausible and continuous throughout a complete action. Figure~\ref{fig:motivation} illustrates this observation with a soccer-kicking sequence. When only a few isolated frames are considered, the player poses, ball positions, and scene content in both the real and generated videos may appear similarly coherent. However, the difference gradually emerges when the same local regions are tracked over time and their successive positions are connected into trajectories. In the real video, the adjustment of body balance, the swing of the kicking leg, and the motion of the ball after contact remain coordinated as the action unfolds. In contrast, an AI-generated video produce visually smooth transitions between adjacent frames, yet struggle to preserve such coordination throughout the complete action.

These limitations suggest that improving the appearance encoder alone provides an unreliable solution. Instead, the detection evidence should be tied more closely to how motion itself evolves throughout an action. We therefore follow stable local regions over time and represent a video as a collection of sparse motion trajectories. Compared with RGB representations, these trajectories retain the continuous evolution of local motion while discarding much of the appearance information that is irrelevant to physical consistency. They are therefore naturally suited to expose motion anomalies that are difficult to identify from individual frames, while requiring neither semantic annotations nor heavy pretrained video backbones.

Based on this idea, we propose \textbf{MotionPhys}, a lightweight and interpretable framework that converts local motion trajectories into physical evidence for AI-generated video detection. To obtain reliable motion observations across videos with different textures and scene contents, MotionPhys combines Shi-Tomasi feature points with regularly sampled grid points and tracks them using pyramidal Lucas-Kanade optical flow. Rather than relying on raw displacement or motion magnitude, we characterize each trajectory in terms of geometric bending, directional change, and temporal smoothness using curvature, angular velocity, and jerk. These motion properties are examined at multiple temporal scales, enabling the framework to retain both short-lived local irregularities and inconsistencies that accumulate throughout an action. Robust statistical aggregation across trajectories and temporal scales further reduces the influence of individual tracking errors. Since the discriminative motion patterns are already encoded in this representation, MotionPhys achieves effective detection using only a lightweight classifier, without relying on a complex video backbone or generator-specific visual artifacts.

Our contributions can be summarized as follows:
\begin{itemize}
\item We investigate AI-generated video detection from the perspective of physical motion consistency. We point out that visual continuity does not necessarily imply real physical motion consistency, and use local optical-flow trajectories as the core observation for detecting implicit motion anomalies in AI-generated videos.

\item We propose \textbf{MotionPhys}, a trajectory-based framework for modeling the physical consistency of local motion over time. Our model captures subtle motion deviations across different temporal ranges and converts them into a compact video-level representation for lightweight and interpretable AI-generated video detection.

\item We evaluate the effectiveness of MotionPhys on public benchmarks including GenBuster++ and AIGVDBench. Experimental results and ablation studies show that MotionPhys captures implicit motion anomalies in AI-generated videos and achieves favorable cross-generator generalization performance.
\end{itemize}

\section{Related Work}

\subsection{AIGC Video Detection Methods}

As the quality of video generation continues to improve, effective AIGC video detection becomes increasingly important. Traditional deepfake detection methods mainly focus on facial cues, such as identity inconsistencies, landmark deformations, head-pose inconsistencies, and local texture artifacts. In recent years, some works have improved generalization ability ability by studying image-level generative artifacts. Tan et al. proposed NPR, which captures structural artifacts introduced by up-sampling operations in generative networks through neighboring pixel relationships~\cite{tan2024rethinking}. Wang et al. found that CNNs trained with proper augmentation can detect various generated images~\cite{wang2020cnn}. Ojha et al. further exploited features from pretrained vision-language models to improve cross-generator generalization~\cite{ojha2023towards}.

For videos generated by diffusion models and large-scale video generation models, recent studies have further extended detection to video-level spatio-temporal modeling. Chen et al. proposed DeMamba, which introduces a Detail Mamba module to model local spatio-temporal inconsistency and builds a million-scale GenVideo benchmark for evaluating AI-generated video detection~\cite{chen2026demamba}. Ma et al. proposed DeCoF, which weakens the influence of spatial artifacts and focuses on frame consistency to detect temporal anomalies in generated videos~\cite{ma2024decof}. In addition, Bai et al. learned spatio-temporal anomalies from both RGB and optical-flow branches~\cite{bai2024ai}. The study \emph{What Matters in Detecting AI-Generated Videos like Sora?} analyzes differences between real and generated videos in terms of appearance, motion, and geometry, while D3 constructs a training-free detector based on second-order central difference features to characterize second-order dynamic shifts in generated videos~\cite{chang2024matters,zheng2025d3}. Recent methods such as NAMP, FOA, BusterX, and BusterX++ also explore AIGC video detection from the perspectives of noise amplification, wavelet-based forensic augmentation, and multimodal large language model-based explanations~\cite{cheng2026revealing,corvi2026seeing,wen2025busterx++}. Beyond these representative approaches, many recent studies have further advanced AIGC video detection and evaluation through diverse methodological and benchmark designs~\cite{feng2026cocovideo,tang2026artifact,li2025skyra,park2025vidguard,hyun2026split}. Although these methods have extended detection cues from appearance artifacts to more specific and interpretable signals, most of them still rely on statistical shifts in frame-level representations. They rarely explicitly analyze whether the continuous motion process of a specific object or local region over time truly follows real-world physical laws.

\subsection{Video Generation}

In recent years, video generation methods based on diffusion models and transformers have made significant progress. The research focus has gradually shifted from improving single-frame visual quality to modeling smooth object motion and scene structure over time. Khachatryan et al. proposed Text2Video-Zero, which extends text-to-image diffusion models to zero-shot text-to-video generation by introducing motion dynamics into latent codes and using cross-frame attention~\cite{clark2023text}. Zhang et al. proposed I2VGen-XL, which improves detail continuity and high-resolution quality in image-to-video generation through image conditioning and cascaded diffusion models~\cite{zhang2023i2vgen}. Blattmann et al. proposed Stable Video Diffusion, emphasizing the importance of large-scale video pretraining, data curation, and high-quality fine-tuning for stable video generation~\cite{blattmann2023stable}. For more complex temporal modeling, Guo et al. proposed AnimateDiff, which learns transferable video motion priors through a plug-and-play motion module~\cite{guo2023animatediff}. Chen et al. proposed SEINE, which uses a randomly masked video diffusion model for short-to-long video transition and prediction~\cite{chen2023seine}.

Subsequently, models such as VideoCrafter, CogVideoX, EasyAnimate, Pyramid Flow, LTX-Video, HunyuanVideo, and Wan further improve visual quality, text alignment, and temporal continuity from different perspectives, including open-domain generation, long-video modeling, diffusion transformers, flow matching, highly compressed latent representations, and large-scale data training~\cite{chen2023videocrafter1,yang2025cogvideox,xu2024easyanimate,jin2025pyramidal,hacohen2024ltx,kong2024hunyuanvideo,wan2025wan}. Although these methods have greatly improved the clarity and temporal coherence of generated videos, their training objectives are still mainly based on distribution matching in feature spaces, without explicitly enforcing real-world physical constraints on local object motion.

\section{Method}

\begin{figure*}[t]
\centering
\includegraphics[width=\textwidth]{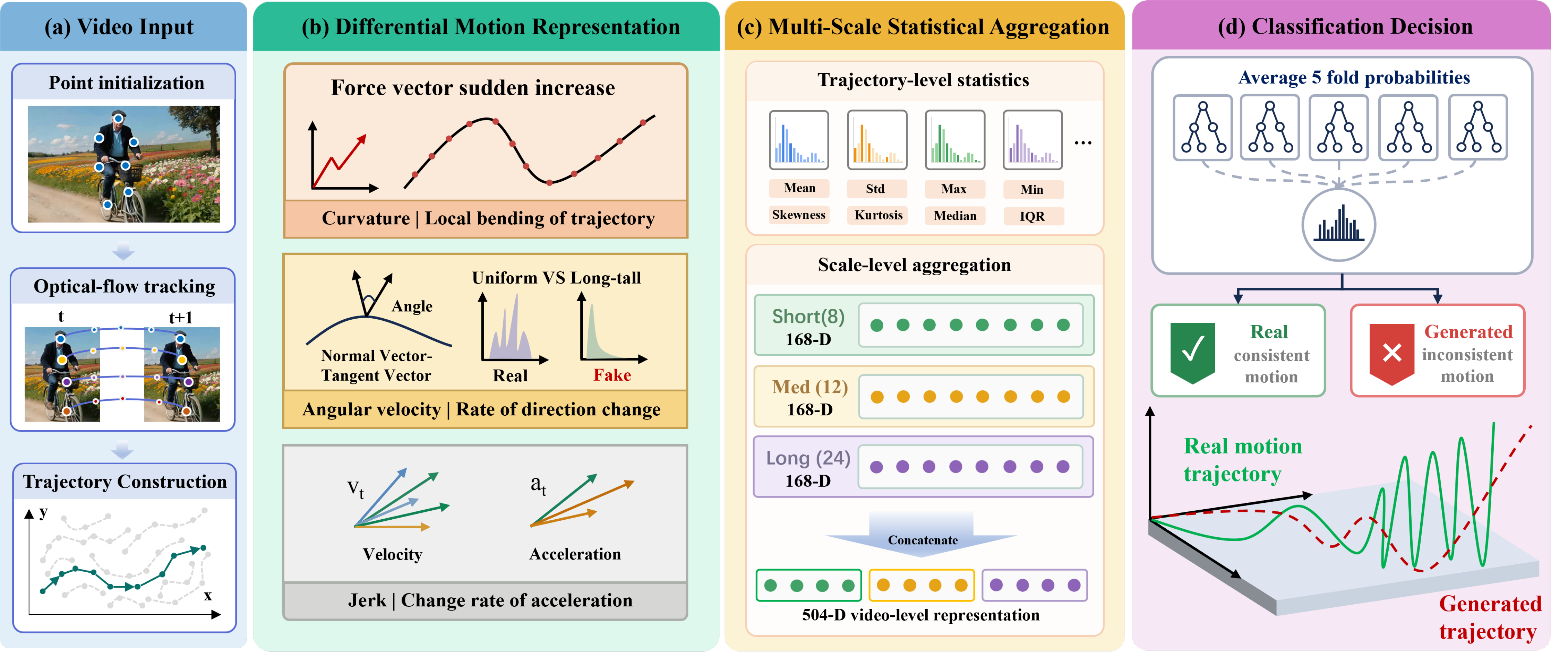}
\caption{Framework of the proposed MotionPhys. (a) Video Input initializes feature points, tracks their inter-frame motion, and connects the tracked positions into motion trajectories. (b) Differential Motion Representation characterizes the trajectories using curvature, angular velocity, and jerk to capture their geometric and temporal evolution. (c) Multi-Scale Statistical Aggregation summarizes the trajectory statistics over 8-, 12-, and 24-frame temporal scales and concatenates them into a 504-dimensional video-level representation. (d) Classification Decision performs the final prediction using a lightweight classifier.}
\label{fig:framework}
\end{figure*}

We aim to develop an interpretable detector that identifies physical motion inconsistencies in AI-generated videos. Figure~\ref{fig:framework} outlines the main components of \textbf{MotionPhys}. MotionPhys tracks local points from sampled video frames to construct motion trajectories at different temporal scales and characterizes their geometric and temporal evolution. The resulting trajectory statistics are aggregated into a video-level physical consistency representation and fed into a lightweight classifier to distinguish real videos from AI-generated ones.

\subsection{Video Input and Trajectory Extraction}

Given an input video $V$, we uniformly sample $N=48$ frames over its entire duration. Tracking points are then initialized by combining Shi--Tomasi corner detection with regular grid sampling. The corner points provide reliable observations in textured regions, while the grid points improve spatial coverage in relatively smooth regions. The initialized points are tracked frame by frame using the pyramidal Lucas--Kanade optical flow algorithm. For each point, the tracker estimates its inter-frame displacement and updates its position in the next frame. Connecting the successfully tracked positions over time produces a two-dimensional motion trajectory on the image plane. Different motion patterns evolve over different temporal durations. Short trajectories are sensitive to transient local changes, whereas longer trajectories describe more persistent motion evolution. We therefore extract trajectories over three temporal scales of 8, 12, and 24 frames.

For each scale $L\in\{8,12,24\}$, we divide the sampled sequence into clips of length $L$ and perform tracking within each clip. The $i$-th trajectory at scale $L$ is represented as
\begin{equation}
\tau_i^L=\{p_{i,t}^L\}_{t=1}^{T_i^L}.
\end{equation}
Each tracked position satisfies
\begin{equation}
p_{i,t}^L\in\mathbb{R}^2,
\end{equation}
where $p_{i,t}^L$ denotes the position of the tracked point at time step $t$, and $T_i^L$ is the valid trajectory length.

\subsection{Differential Motion Representation}

A motion trajectory can be viewed as a discrete temporal curve whose geometric and dynamic evolution reflects the underlying motion process. Based on this view, MotionPhys characterizes each trajectory using curvature, angular velocity, and jerk, which provide complementary descriptions of trajectory bending, directional evolution, and high-order temporal smoothness.

\subsubsection{Kinematic Quantities}

For a trajectory $\tau=\{p_t\}_{t=1}^{T}$, the velocity vector is computed using the first-order temporal difference:
\begin{equation}
v_t=p_{t+1}-p_t.
\end{equation}
The acceleration vector is then obtained by differentiating the velocity:
\begin{equation}
a_t=v_{t+1}-v_t.
\end{equation}
These kinematic quantities form the basis for constructing the subsequent geometric and dynamic descriptors.

\subsubsection{Curvature}

Curvature measures the local bending of a motion trajectory. For a discrete two-dimensional trajectory, it is computed from the velocity and acceleration vectors as
\begin{equation}
\kappa_t=\frac{\left|v_t^x a_t^y-v_t^y a_t^x\right|}{\|v_t\|_2^3+\epsilon},
\end{equation}
where $v_t^x$ and $v_t^y$ denote the horizontal and vertical components of $v_t$, $a_t^x$ and $a_t^y$ are the corresponding components of $a_t$, and $\epsilon$ is a small constant for numerical stability.

Curvature captures the geometric evolution of a local motion path. Natural motion generally exhibits gradual trajectory bending under physical and structural constraints, whereas generated videos may contain over-smoothed segments or abrupt local changes that produce abnormal curvature patterns.

\subsubsection{Angular Velocity}

Curvature describes the geometric shape of a trajectory, but trajectories following similar paths may still differ in how their directions evolve over time. We first normalize the instantaneous motion direction as
\begin{equation}
\hat{v}_t=\frac{v_t}{\|v_t\|_2+\epsilon}.
\end{equation}
The angular change between adjacent time steps is then computed as
\begin{equation}
\omega_t=\arccos\!\left(\mathrm{clip}\left(\hat{v}_t^\top\hat{v}_{t+1},-1,1\right)\right).
\end{equation}
With the temporal interval normalized to one frame, $\omega_t$ represents the magnitude of angular velocity in radians per frame. The clipping operation prevents numerical errors caused by values slightly outside the valid range of the $\arccos$ function.

Angular velocity captures directional changes during trajectory traversal and complements the geometric information provided by curvature. Abnormal angular velocity patterns may indicate unstable or physically inconsistent turning behavior.

\subsubsection{Jerk}

Jerk describes high-order motion smoothness and is sensitive to abrupt changes in frame-to-frame motion. For a discrete trajectory, the jerk vector is computed as
\begin{equation}
j_t=a_{t+1}-a_t=p_{t+3}-3p_{t+2}+3p_{t+1}-p_t.
\end{equation}
Its magnitude is defined as
\begin{equation}
\rho_t=\|j_t\|_2.
\end{equation}
The jerk magnitude captures discontinuities in acceleration and provides complementary evidence for identifying unnatural local motion transitions.

\subsubsection{Multi-Scale Statistical Aggregation}

The curvature, angular velocity, and jerk values computed along a trajectory form variable-length temporal sequences. To obtain a fixed-dimensional description, each physical quantity sequence $q$ is summarized using eight statistical measures:
\begin{equation}
\phi(q)=\left[\mu(q),\sigma(q),q_{\min},q_{\max},\gamma_1(q),\gamma_2(q),\tilde{q},\mathrm{IQR}(q)\right],
\end{equation}
where $\mu$ and $\sigma$ denote the mean and standard deviation, $q_{\min}$ and $q_{\max}$ are the minimum and maximum values, $\gamma_1$ and $\gamma_2$ denote skewness and kurtosis, $\tilde{q}$ is the median, and $\mathrm{IQR}$ is the interquartile range.

For the $i$-th trajectory at temporal scale $L$, we apply $\phi(\cdot)$ to its curvature sequence $\kappa_i^L$, angular velocity sequence $\omega_i^L$, and jerk sequence $\rho_i^L$. The trajectory-level feature is defined as
\begin{equation}
f_i^L=\left[\phi(\kappa_i^L);\phi(\omega_i^L);\phi(\rho_i^L)\right].
\end{equation}
This produces a 24-dimensional descriptor:
\begin{equation}
f_i^L\in\mathbb{R}^{24}.
\end{equation}

Because the number of valid trajectories varies across videos, the trajectory-level features must be further aggregated into a fixed-dimensional scale-level representation. For temporal scale $L$, the features of all $M_L$ valid trajectories are stacked into
\begin{equation}
F^L=\left[f_1^L,f_2^L,\ldots,f_{M_L}^L\right]^\top.
\end{equation}
The resulting feature matrix satisfies
\begin{equation}
F^L\in\mathbb{R}^{M_L\times24}.
\end{equation}

For each feature dimension $d\in\{1,\ldots,24\}$, we compute seven statistics across the corresponding values from all trajectories:
\begin{equation}
\psi(F_{:,d}^L)=\left[\mu_d,\sigma_d,f_{d,\min},f_{d,\max},\tilde{f}_d,Q_{25,d},Q_{75,d}\right],
\end{equation}
where $\mu_d$ and $\sigma_d$ are the mean and standard deviation, $f_{d,\min}$ and $f_{d,\max}$ are the minimum and maximum values, $\tilde{f}_d$ is the median, and $Q_{25,d}$ and $Q_{75,d}$ are the 25th and 75th percentiles.

The scale-level representation is obtained by concatenating the aggregation results of all 24 feature dimensions:
\begin{equation}
z^L=\left[\psi(F_{:,1}^L);\psi(F_{:,2}^L);\ldots;\psi(F_{:,24}^L)\right].
\end{equation}
Since each feature dimension is summarized by seven statistics, the scale-level representation has
\begin{equation}
z^L\in\mathbb{R}^{168}.
\end{equation}

Finally, the representations from the three temporal scales are concatenated as
\begin{equation}
z=\left[z^8;z^{12};z^{24}\right].
\end{equation}
The resulting video-level physical consistency representation satisfies
\begin{equation}
z\in\mathbb{R}^{504}.
\end{equation}
This multi-scale representation preserves motion inconsistencies occurring over different temporal ranges while reducing the influence of individual trajectories and local tracking errors.




\begin{table}[t]
\centering
\small
\renewcommand{\arraystretch}{1.08}
\setlength{\tabcolsep}{3pt}

\begin{tabular*}{\columnwidth}{@{\extracolsep{\fill}}cccc@{}}
\hline
Method & Real & Fake & Overall \\
\hline
Keye-VL-8B
& \textbf{95.70} & 11.80 & 53.75 \\

MiniCPM-o 2.6
& 78.40 & 29.80 & 54.10 \\

Qwen2.5-Omni-7B
& 81.40 & 31.40 & 56.40 \\

Qwen2.5-VL-7B
& 92.60 & 27.70 & 60.15 \\

InternVL3-8B
& 80.10 & 41.30 & 60.70 \\

MiMo-VL-7B-RL
& 86.10 & 42.70 & 64.40 \\

BusterX
& 86.40 & 53.10 & 69.75 \\

BusterX++
& \underline{95.30} & \underline{57.90} & \underline{76.60} \\

\textbf{MotionPhys(Ours)}
& 84.20
& \textbf{72.80}
& \textbf{78.50} \\
\hline
\end{tabular*}
\caption{Performance comparison on GenBuster++. Following the official evaluation protocol, MotionPhys is trained on GenBuster and evaluated on GenBuster++. Real and Fake denote the class-wise accuracies on real and AI-generated videos, respectively, while Overall reports the accuracy over the entire test set. All results are reported in percentages. Best in each column is bold, second best is underlined.}
\label{tab:genbuster-main}
\end{table}

\subsection{Classification Decision}

\begin{table*}[t]
\centering
\fontsize{8}{9}\selectfont
\setlength{\tabcolsep}{0pt}
\renewcommand{\arraystretch}{1.12}
\newcommand{\aigvdhead}[2][1]{%
  \raisebox{-0.5\height}{\scalebox{#1}[1]{\shortstack[c]{#2}}}%
}
\begin{tabular*}{\textwidth}{@{\extracolsep{\fill}}c*{20}{c}c@{}}
\toprule
\multirow{2}{*}{Method}
& \multicolumn{6}{c}{I2V}
& \multicolumn{12}{c}{T2V}
& \multicolumn{2}{c}{V2V}
& \multirow{2}{*}{Avg} \\
\cmidrule(lr){2-7}\cmidrule(lr){8-19}\cmidrule(lr){20-21}
& \aigvdhead[0.86]{Easy\\Animate}
& \aigvdhead[0.86]{LTX}
& \aigvdhead[0.84]{Pyramid\\Flow}
& \aigvdhead[0.90]{SEINE}
& \aigvdhead[0.86]{SVD}
& \aigvdhead[0.86]{Video\\Crafter}
& \aigvdhead[0.86]{Acc\\Video}
& \aigvdhead[0.84]{Animate\\Diff}
& \aigvdhead[0.74]{CogVideo\\X1.5}
& \aigvdhead[0.86]{Easy\\Animate}
& \aigvdhead[0.80]{Hunyuan}
& \aigvdhead[0.86]{IPOC}
& \aigvdhead[0.86]{LTX}
& \aigvdhead[0.86]{Open\\Sora}
& \aigvdhead[0.84]{Pyramid\\Flow}
& \aigvdhead[0.86]{Rep\\Video}
& \aigvdhead[0.86]{Video\\Crafter}
& \aigvdhead[0.86]{Wan\\2.1}
& \aigvdhead[0.74]{CogVideo\\X1.5}
& \aigvdhead[0.86]{LTX}
& \\
\midrule
\multicolumn{22}{c}{\textbf{Video Classification Models}} \\
\midrule
MViTv2
& 52.3 & 67.7 & 65.9 & 72.6 & 64.6 & 70.5 & 64.9 & 72.8 & 63.6 & 60.7 & 69.1 & 59.3 & 69.5 & 100.0 & 70.0 & 56.3 & 68.6 & 48.7 & 62.2 & 64.9 & 66.2 \\

UniFormer
& 48.5 & 60.1 & 66.4 & 44.4 & 32.0 & 68.8 & \textbf{85.0} & \underline{80.5} & 67.6 & 66.0 & 84.8 & 72.8 & 52.9 & 89.5 & 84.9 & 66.4 & 83.7 & 55.4 & 75.1 & 69.4 & 67.7 \\

VideoSwin
& 60.4 & 70.9 & 65.6 & 90.7 & 85.1 & 77.1 & 62.9 & 73.7 & 65.6 & 63.0 & 60.8 & 63.7 & 70.1 & 100.0 & 63.9 & 60.2 & 76.3 & 55.0 & 57.5 & 66.5 & 69.5 \\

VideoMAE
& 57.5 & 69.1 & 77.0 & 69.4 & 67.0 & 65.4 & 78.7 & 66.4 & 75.3 & 66.9 & 81.0 & 83.6 & 75.0 & 98.9 & 92.7 & 74.4 & 86.7 & 62.7 & 71.3 & 71.5 & 74.5 \\

TSM
& \underline{79.4} & 70.4 & 85.9 & 89.7 & 93.8 & 84.3 & 60.1 & 58.4 & 68.5 & 77.6 & 70.6 & 72.7 & 80.1 & 100.0 & 87.0 & 68.9 & 94.5 & 56.0 & 68.9 & 66.7 & 76.7 \\

SlowFast
& 63.9 & 71.6 & 83.8 & 74.0 & 72.3 & 68.4 & 82.7 & 71.0 & 80.2 & 74.9 & \underline{83.5} & 89.4 & 76.2 & \underline{99.8} & 92.8 & 79.9 & 91.3 & 65.9 & \underline{78.9} & \textbf{77.5} & 78.9 \\

TimeSformer
& 66.3 & 70.4 & 80.3 & 82.1 & 88.3 & 65.4 & \underline{83.9} & 76.3 & \textbf{84.6} & 84.3 & \textbf{84.8} & \textbf{91.3} & \underline{85.5} & 100.0 & 96.2 & \textbf{85.1} & 96.4 & \underline{73.7} & 69.6 & 65.6 & 81.5 \\

I3D
& \textbf{81.0} & \textbf{81.7} & \textbf{94.0} & \textbf{96.7} & \underline{96.1} & \underline{84.8} & 66.6 & 46.8 & \underline{83.5} & 78.4 & 75.5 & \underline{90.2} & \textbf{94.6} & 100.0 & \underline{96.8} & \underline{80.3} & 91.5 & 65.3 & \textbf{79.7} & \underline{76.4} & 83.0 \\

UniFormerV2
& 73.9 & \underline{77.6} & 77.1 & 82.4 & 94.3 & 82.3 & 82.5 & \textbf{86.6} & 83.0 & \underline{84.4} & 81.5 & 84.5 & 82.6 & 100.0 & 93.0 & 79.9 & \underline{97.1} & \textbf{76.4} & 73.9 & 72.7 & \underline{83.3} \\

X3D
& \textbf{91.0} & 68.4 & \underline{92.8} & \underline{96.1} & \textbf{97.5} & \textbf{96.6} & 79.5 & 67.8 & 76.9 & \textbf{93.8} & 80.4 & 84.3 & 84.2 & \textbf{100.0} & \textbf{97.1} & 78.8 & \textbf{97.5} & 63.6 & 73.5 & 61.7 & \textbf{84.1} \\

\midrule
\multicolumn{22}{c}{\textbf{AI-Generated Video Detection Models}} \\
\midrule
DeMamba
& \textbf{76.4} & \underline{77.6} & \textbf{92.8} & \underline{94.8} & \underline{98.9} & \underline{52.0} & 76.8 & 64.8 & 82.2 & 76.6 & 73.8 & \underline{90.9} & \underline{85.3} & \textbf{100.0} & \textbf{98.9} & \underline{78.7} & 88.0 & \underline{66.4} & 67.5 & 77.5 & 81.0 \\

DeCoF
& 70.6 & \textbf{81.5} & 83.7 & 85.7 & 93.1 & 45.2 & \underline{79.5} & \underline{68.1} & \textbf{90.7} & \underline{78.0} & \underline{79.8} & \textbf{94.6} & \textbf{87.7} & 99.9 & \underline{96.6} & \textbf{89.5} & \underline{90.8} & \textbf{72.9} & \underline{81.8} & \textbf{81.3} & \underline{82.6} \\

\textbf{MotionPhys(Ours)}
& \underline{74.7} & 74.4 & \underline{85.2} & \textbf{99.4} & \textbf{99.4} & \textbf{98.8} & \textbf{79.9} & \textbf{99.2} & \underline{83.4} & \textbf{78.4} & \textbf{82.6} & 76.2 & 77.3 & \underline{99.9} & 92.1 & 73.6 & \textbf{99.5} & 60.2 & \textbf{84.6} & \underline{78.3} & \textbf{84.8} \\
\bottomrule
\end{tabular*}
\caption{Performance comparison on AIGVDBench under the official cross-generator evaluation protocol, where the detector is trained on Open-Sora and evaluated on all other unseen generators. Each entry reports AUC, Avg is the mean over all listed models.}
\label{tab:aigvd-main}
\end{table*}

After multi-scale trajectory distribution aggregation, each video is represented as a 504-dimensional physical consistency feature vector. Unlike end-to-end video detection networks, the main discriminative information of MotionPhys comes from explicitly constructed trajectory-level physical representations, rather than visual features implicitly learned from RGB frames by large-scale neural networks. Therefore, the classifier does not need to perform complex spatio-temporal representation learning. Instead, it only needs to model the nonlinear interactions among different dimensions in the physical statistical feature space.

The resulting representation is compatible with a broad range of conventional classifiers and does not rely on a particular decision model. In our implementation, we adopt LightGBM as the default classifier. Since the input features have explicit physical meanings, the classification results can be further traced back to different physical dimensions, such as geometric anomalies in trajectories, anomalies in turning dynamics, and anomalies in higher-order motion smoothness. This makes MotionPhys more interpretable than end-to-end visual classifiers.

\section{Experiments}

\subsection{Experimental Setup}

\subsubsection{Datasets}

We evaluate MotionPhys on two public benchmarks for AI-generated video detection, namely GenBuster++ and AIGVDBench. GenBuster++ is a carefully curated cross-modal benchmark containing high-quality video samples from diverse real-world scenarios.

AIGVDBench is a large-scale benchmark comprising over 440K videos generated by 31 recent video generation models. It covers text-to-video, image-to-video, and video-to-video generation paradigms and includes diverse semantic content and real-world scenarios, enabling a comprehensive evaluation of detector generalization across generation settings.

\subsubsection{State-of-the-Art Methods.}
We compare MotionPhys with a broad range of representative methods on GenBuster++ and AIGVDBench. On GenBuster++, the compared methods include recent multimodal large language models Keye-VL-8B~\cite{team2025kwai}, MiniCPM-o 2.6~\cite{yao2024minicpm}, Qwen2.5-Omni-7B~\cite{xu2025qwen25omnitechnicalreport}, Qwen2.5-VL-7B~\cite{bai2025qwen25vltechnicalreport}, InternVL3-8B~\cite{zhu2025internvl3}, and MiMo-VL-7B-RL~\cite{coreteam2025mimovltechnicalreport}, together with the recent video forgery detectors BusterX~\cite{wen2025busterx} and BusterX++~\cite{wen2025busterx++}. On AIGVDBench, we evaluate commonly used video classification models, including MViTv2~\cite{li2022mvitv2}, UniFormer~\cite{li2022uniformer}, VideoSwin~\cite{liu2022video}, VideoMAE~\cite{tong2022videomae}, TSM~\cite{lin2019tsm}, SlowFast~\cite{feichtenhofer2019slowfast}, TimeSformer~\cite{bertasius2021space}, I3D~\cite{carreira2017quo}, UniFormerV2~\cite{li2022uniformerv2}, and X3D~\cite{feichtenhofer2020x3d}. We also compare with the recent AI-generated video detection methods DeMamba~\cite{chen2026demamba} and DeCoF~\cite{ma2025detecting}.

\subsubsection{Evaluation Metrics}

To comprehensively evaluate detection performance, we report accuracy and AUC. In the tables, all numerical values are reported in percentages.

\subsection{Main Results}

To validate the effectiveness of the proposed MotionPhys method, we compare it with existing detection methods on GenBuster++ and AIGVDBench. Table~\ref{tab:genbuster-main} reports the results on GenBuster++. MotionPhys achieves an overall accuracy of 78.5\%, outperforming all compared methods. Most MLLM-based methods perform reasonably well on real videos but suffer from a significant accuracy drop on fake videos, indicating their limited ability to identify AI-generated video content. In contrast, MotionPhys achieves 84.2\% and 72.8\% accuracy on real and fake videos, respectively, maintaining a more balanced performance. This suggests that physical motion features are more robust to content appearance and can better capture intrinsic motion abnormalities in generated videos.

We further evaluate MotionPhys on AIGVDBench. Specifically, the detector is trained on Open-Sora and tested on videos generated by other models, including I2V, V2V, and T2V generators. Table~\ref{tab:aigvd-main} reports the performance across different generation models. Compared with existing video backbones and forgery detectors, MotionPhys achieves competitive performance on most subsets, showing that the proposed physical trajectory features can capture motion abnormalities across different generation paradigms.

\subsection{Ablation Studies}

To better evaluate our method, we conduct multiple ablation and robustness studies. All ablation experiments are conducted on GenBuster++.

\subsubsection{Contribution of Physical Features}

To analyze the role of different physical features, we train detectors using each individual feature group, each pairwise combination, and all three feature groups. The results are shown in Table~\ref{tab:ablation-features}.

\begin{figure*}[t]
    \centering
    \includegraphics[width=\textwidth]{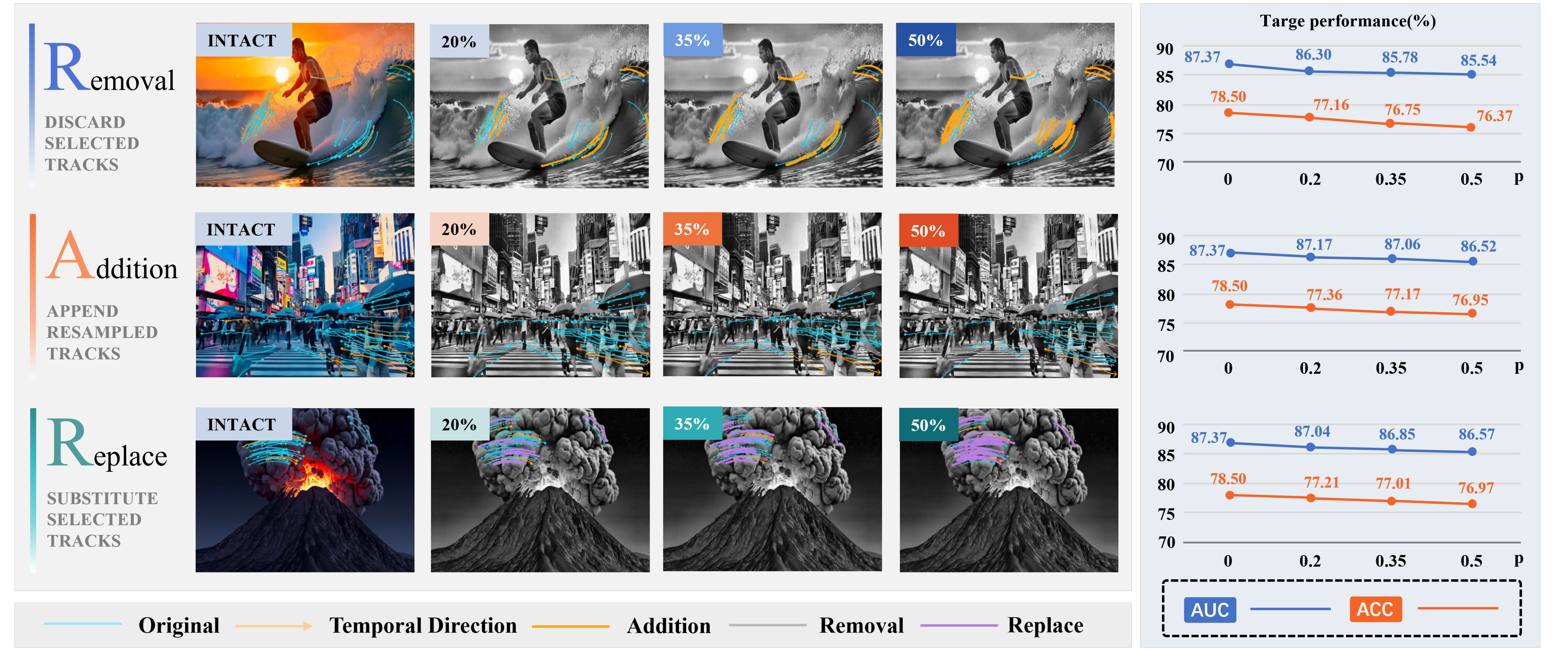}
    \caption{Robustness of MotionPhys to trajectory-set perturbations on GenBuster++. The left side illustrates the changes in the trajectory set under random removal, random addition, and random replacement, while the right side reports AUC and ACC from the clean setting to a perturbation ratio of 50\%.}
    \label{fig:noise-robustness}
\end{figure*}

\begin{table}[t]
\centering

\small
\setlength{\tabcolsep}{4pt}
\renewcommand{\arraystretch}{1.16}
\begin{tabular}{C{0.45cm} C{1.25cm} C{1.25cm} C{1.25cm} C{0.8cm} C{0.8cm}}
\toprule
No. & Curvature & \shortstack{Angular\\Velocity} & Jerk & ACC & AUC \\
\midrule
1 & \cmark & \xmark & \xmark & 74.10 & 81.20 \\
2 & \xmark & \cmark & \xmark & 74.15 & 80.40 \\
3 & \xmark & \xmark & \cmark & 73.60 & 80.01 \\
\midrule
4 & \cmark & \cmark & \xmark & 77.25 & 85.91 \\
5 & \cmark & \xmark & \cmark & 75.80 & 84.50 \\
6 & \xmark & \cmark & \cmark & 76.00 & 84.27 \\
\midrule
7 & \cmark & \cmark & \cmark & \textbf{78.50} & \textbf{87.37} \\
\bottomrule
\end{tabular}

\caption{Ablation of physical feature combinations on GenBuster++. ACC and AUC are reported for individual features, pairwise combinations, and the complete representation.}
\label{tab:ablation-features}
\end{table}

\subsubsection{Analysis of Temporal Scales}
We further compare different temporal-scale configurations, with the results shown in Table~\ref{tab:ablation-scales}. Combining multiple scales further improves performance, and the full three-scale configuration achieves the highest ACC and AUC. These ablation studies demonstrate the effectiveness and robustness of the proposed hierarchical motion differential geometry representation and multi-scale trajectory aggregation strategy. The three feature groups in this representation are complementary, and the multi-scale design jointly captures local details and global motion trends, forming a strong physically grounded representation for AI-generated video detection. and global motion trends, forming a strong physical representation for AI-generated video detection.

\begin{table}[t]
\centering

\small
\setlength{\tabcolsep}{4.5pt}
\renewcommand{\arraystretch}{1.12}
\begin{tabular}{lccccc}
\toprule
Setting & Scales & Frames & Dim. & ACC & AUC \\
\midrule
\multirow{3}{*}{Single}
& $S$ & 8  & 168 & 76.35 & 84.17 \\
& $M$ & 12 & 168 & 76.65 & 85.49 \\
& $L$ & 24 & 168 & 76.55 & 85.43 \\
\midrule
\multirow{3}{*}{Pairwise}
& $S{+}M$ & 8, 12  & 336 & 76.60 & 86.16 \\
& $S{+}L$ & 8, 24  & 336 & 77.20 & 86.70 \\
& $M{+}L$ & 12, 24 & 336 & 77.50 & 86.88 \\
\midrule
Full
& $S{+}M{+}L$ & 8, 12, 24 & 504 & \textbf{78.50} & \textbf{87.37} \\
\bottomrule
\end{tabular}

\caption{Ablation study of temporal-scale configurations on GenBuster++. $S$, $M$, and $L$ denote trajectories spanning 8, 12, and 24 frames, respectively.}
\label{tab:ablation-scales}
\end{table}

\subsubsection{Classifier Comparison}

As shown in Table~\ref{tab:classifier-comparison}, we apply different classifiers to the extracted motion features and compare their performance. Linear classifiers obtain relatively lower results, whereas conventional nonlinear classifiers provide substantial improvements. This indicates that the extracted representation already contains discriminative physical information, while effective classification requires modeling nonlinear interactions among different trajectory statistics. Several lightweight classifiers achieve strong performance, demonstrating that MotionPhys does not depend on a specific decision model.

\begin{table}[t]
\centering

\small
\setlength{\tabcolsep}{5.5pt}
\renewcommand{\arraystretch}{1.08}
\begin{tabular}{lcc}
\toprule
Classifier & AUC & ACC \\
\midrule
Logistic regression & 73.23 & 65.75 \\
Linear SVM           & 73.16 & 66.20 \\
RBF-SVM              & 81.55 & 72.20 \\
$k$-NN               & 81.97 & 72.95 \\
Decision tree        & 82.41 & 73.65 \\
Random forest        & 83.37 & 74.35 \\
ExtraTrees           & 84.75 & 75.40 \\
HistGradientBoosting & 85.83 & 76.55 \\
MLP                   & 86.54 & 77.20 \\
LightGBM              & \textbf{87.37} & \textbf{78.50} \\
\bottomrule
\end{tabular}

\caption{
Comparing representative classifiers on GenBuster++ with identical motion representation, reporting ACC and AUC.
}
\label{tab:classifier-comparison}
\end{table}

\subsection{Robustness to Trajectory Perturbations}
We further investigate how perturbations to the trajectory set affect MotionPhys. Starting from the clean trajectory set, we progressively modify its size and composition through random removal, random addition, and random replacement. Random removal and addition assess the sensitivity of the detector to changes in the number of available trajectories, while random replacement preserves the trajectory count and examines whether the prediction depends on specific trajectories. As shown in Figure~\ref{fig:noise-robustness}, both AUC and ACC vary only slightly as the perturbation ratio increases from 0\% to 50\%. The stable performance across all three settings indicates that the video-level decision is primarily determined by the collective distribution and overall trends of physical motion patterns across the trajectory set rather than by individual trajectories.

\section{Conclusion}

In this paper, we introduce motion differential geometry into AI-generated video detection by deriving physical consistency features from local optical-flow trajectories. Our systematic analysis reveals that, despite their high visual quality and temporal continuity, current AI-generated videos still exhibit systematic deviations from real videos in object motion trajectories. Based on this observation, we propose \textbf{MotionPhys}, a lightweight and interpretable detection framework built upon sparse optical-flow trajectories. By connecting video content with real-world physical motion regularities, MotionPhys provides an intuitive approach for understanding and distinguishing real videos from AI-generated videos.

\subsubsection{Limitations and Future Work.}
MotionPhys primarily characterizes two-dimensional motion patterns on the image plane and does not explicitly model richer physical constraints such as depth variations, object interactions, and contact relationships. Future research could incorporate additional physical consistency cues while preserving the interpretability of the framework, thereby improving its adaptability to complex motion scenarios. We believe that this work will encourage further research on physical artifacts in generated videos and contribute to the development of more generalizable and interpretable detection techniques.

\bibliography{aaai2027}

\end{document}